\documentclass[]{svproc}
\usepackage[T1]{fontenc}
\usepackage[utf8]{inputenc}
\usepackage{mathtools}
\usepackage{url}

\begin{document}
\mainmatter              
\title{Shared Autonomy in Web-based Human Robot Interaction}
%
%
\author{Yug Ajmera, Arshad Javed}
\authorrunning{Yug Ajmera et al.} 
%

%
\institute{Mechanical Department, Birla Institute of Technology and Science, Pilani, Hyderabad, India\\
\email{\{f20170644,arshad\}@hyderabad.bits-pilani.ac.in}}

\maketitle              

\begin{abstract}
In this paper, we aim to achieve a human-robot work balance by implementing shared autonomy through a web interface. Shared autonomy integrates user input with the autonomous capabilities of the robot and therefore increases the overall performance of the robot. Presenting only the relevant information to the user on the web page lowers the cognitive load of the operator. Through our web interface, we provide a mechanism for the operator to directly interact using the displayed information by applying a point-and-click paradigm. Further, we present our idea to employ a human-robot mutual adaptation in a shared autonomy setting through our web interface for effective team collaboration.
\end{abstract}

\section{Introduction}
There has been an increase in the number of applications for robot teleoperation including military \cite{kot2018application}, industrial \cite{korpela2015applied}, surveillance \cite{lopez2013watchbot}, telepresence \cite{8948789} and remote experimentation \cite{pitzer2012pr2}. Improving operator efficiency and ensuring the safe navigation of robots is of utmost importance. Studies show that human-robot joint problem solving results in safe and effective task execution \cite{music2017control}. Humans are better at reasoning and creativity whereas robots are better at carrying out a particular task precisely and repeatedly. Therefore, combining robot capabilities with human skills results in an enhanced human-robot interaction. 

Having remote access to robots through user-friendly interfaces is very important for effective human-robot interaction. Different methods for the control of mobile robots are being developed and tested. Gomez has developed a GUI for teleoperation of robots for teaching purposes \cite{gomez2016learning}. The GUI is created using Qt creator which reduces the usability of the system in comparison to a web-based interface. Lankenau's telepresence system called Virtour provides remote access to wheeled robots through the website \cite{lankenau2016virtour}. The tour leader controls the robot whereas the guest robots follow it. This provides relatively limited autonomy since the control lies solely in the hands of an operator. Our web-interface is unique in a way that it allows the users
point-and-click navigation to arbitrary locations in unknown environments. The interface described in \cite{birkenkampf2014knowledge}, implements shared autonomy but is limited to a tablet computer. We have tried to overcome these typical shortcomings like poor accessibility and usability through our web-based interface. Such an interface would enable a simpler control of the robot for novices and experts alike. This allows users to control robots within their home or workplace through any web-enabled devices. The web clients are created using modern web standards hence it does not require users to download any extra software in order to use it. Furthermore, we have tried to display all the visualization data on the web page. This makes it convenient for users to perform the entire navigation process by just using the web interface.

The objective of this research is to develop an intuitive  web interface for robot control using various levels of autonomy. The interface is initially built and tested on the three-wheeled telepresence robot of our lab. It is an open-ended design and can be extended to various use cases. Further, we describe our idea to integrate a bounded-memory adaptation model (BAM) of the human teammate into a partially observable stochastic process to enable a robot to adapt to a human. Studies show that this retains a high-level of user trust in robots and significantly improves human-robot team performance \cite{nikolaidis2017human}.

\section{Web-interface}
\subsection{Software Infrastructure}
ROS (Robot Operating System) is used as a back-end system for most robots nowadays. It is an open-source middle-ware platform that provides several useful libraries and tools to develop robot applications. The telepresence robot of our lab is based on ROS. We have implemented the ROS Navigation stack to perform autonomous navigation.

\begin{figure}[h]
  \centering
  \includegraphics[width=\linewidth]{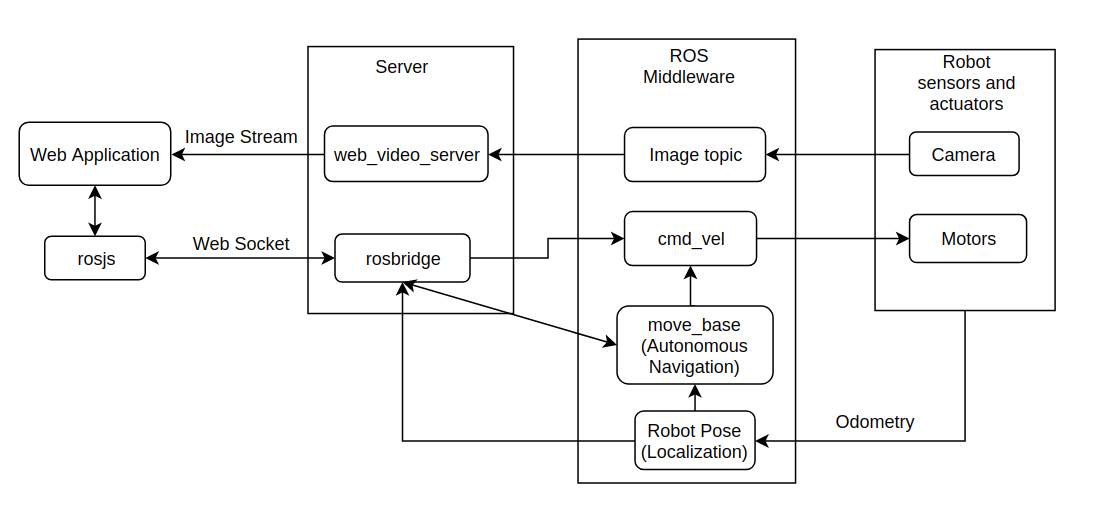}
  \caption{System flow of the web interface}
\end{figure}

The graphical user interface is created using HTML, CSS, and Javascript. Apart from this, Robot web tools \cite{toris2015robot} are used to provide a web connection to the ROS system. For interaction between ROS and web page Javascript requests, we have used rosbridge \cite{crick2017rosbridge}. The user’s activities on the web page are interfaced as JavaScript Object Notation(JSON) commands which are then converted to ROS commands. Roslibjs \cite{osentoski2011robots} is a standard ROS javascript library that connects rosbridge and the web application. It enables the interface to publish and subscribe to ROS topics. Through the use of web sockets, rosbridge and roslibjs can be readily used with modern web browsers without any installation. This makes it an ideal platform for us. Furthermore, rosbridge provides the feature of data logging. Analyzing the logged data and correcting the errors can increase the efficiency drastically.

For hosting our web page, we employed the roswww package which provides an HTTP web server at a specified port. Hence the user can access the web page as long as they are connected to the wi-fi shared by the robot. Finally, the web\_video\_server package is used to display the live video feed from the camera of the robot to the web page. It streams the images through an image topic in ROS via HTTP.

\subsection{User Interaction}
\begin{figure}[h]
  \centering
  \includegraphics[width=\linewidth]{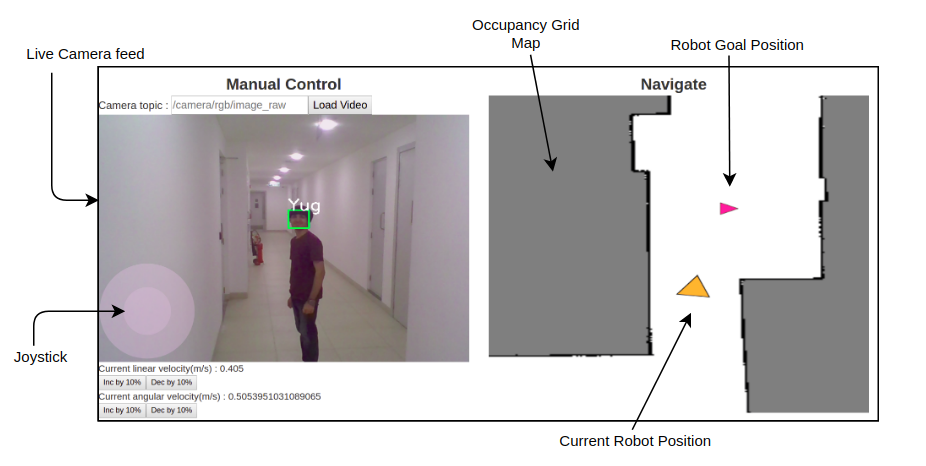}
  \caption{The graphical user web interface}
\end{figure}

The web interface (shown in Fig 2.) is divided into two parts: Manual teleoperation and autonomous navigation. In manual control, the user is provided with an on-screen touch-capable joystick and the live video feed from the camera of the robot. The extent of pull determines the fraction of the maximum velocity. Using the orientation, corresponding linear and angular velocities are calculated. These are then published in the velocity topic as a geometry/Twist message. The current maximum linear and angular speeds of the robot are displayed. Two buttons each are provided for increasing or decreasing these velocities by ten percent of the current velocity. By default, the camera topic is set as /camera/rgb/image\_raw. The user can change it by typing a different topic name in the space provided. On the click of the “Load Video” button, the video feed loads.

A joystick is used instead of buttons because it employs a game-based strategy to teleoperate robots that are intuitive even for non-expert users. For better teleoperation, we need to enhance the information provided to the user. Hence, we have provided real-time video data on the web page. This provides a robot-centered perspective to the user and it replicates the environment which the user would perceive in place of the robot.

The autonomous navigation section allows the users point-and-click navigation to arbitrary locations. The current position of the robot is displayed as a yellow pulsating arrow on the map. The user can give a goal position and orientation by clicking on an arbitrary point on the map. The goal is marked by a red arrow. This is sent to the move\_base node which plans a safe path to the goal. In that instance, the robot starts moving autonomously towards the goal position.

\section{Shared Autonomy}
In the case of full control or teleoperation method, the user has to manually guide the robot through the desired path using the joystick. Because the user has to constantly be in control of the robot as well as be aware of the surroundings of the robot until the execution of the task, this method is cumbersome. On the other hand, in the case of full autonomy, there is no involvement of the user once the goal position is marked. The robot autonomously navigates through the obstacles and reaches the goal position through the shortest route. If the user-intended path is not the shortest path then the robot fails to meet the expectations of the user. Therefore, shared autonomy is necessary for an efficient human-robot interaction. 

The web application presented in this paper, provides control over the robot to the user, while simultaneously using the existing autonomous navigation capabilities and obstacle avoidance to ensure safety and correct operation. It enables varied autonomy of the robot during the execution of tasks. This changes the level of user involvement in carrying out the tasks. When the user marks the goal position on the map, the move\_base ROS node calculates the shortest path using Dijkstra's algorithm and the robot starts following the path. At any point, if the user feels that the robot is malfunctioning, or if the user wants the robot to follow a different path to reach the goal, the user can override the control using the joystick, and guide the robot to that path. A ROS service then republishes the original goal to move\_base ROS node and the robot re-plans its trajectory.

\subsection{Mutual Adaptation}
In this section, we propose an alternative method to employ shared autonomy using mutual adaptation through our web interface. In order to implement mutual adaptation, the robot should not only suggest efficient strategies which may be unknown to the user but it should also comply with human's decision in order to gain his trust. We assume that the robot knows the optimal goal for the task: to reach the goal position. Consider a situation where the robot has two choices to avoid the obstacle: it can take the right or the left path. Taking the right path is a better choice, for instance, because the left path is too long, or because the robot has less uncertainty about the right part of the map. Intuitively, if the human insists on the left path, the robot should comply; failing to do so can have a negative effect on the user’s trust in the robot, which may lead to the disuse of the system. 

We formulate the problem with world state $x_{world} \in X_{world}$, robot action $a_r \in A_r$ and human action  $a_h \in A_h$. The goal is assumed to be among a discrete set of goals $g \in G$. The state transition function is defined as $T: X_{world} \times A_r \times A_h \xrightarrow{} \prod(X_{world})$. The human actions $a_h \in A_h$ are read through the inputs of the web interface and hence does not affect the world state. The state transition function reduces to $T: X_r \times A_r \xrightarrow{} X_r$. 

The Boundary-memory Adaptation Model (BAM) simplifies the problem by limiting the history length to k steps. Based on a history of k steps, we compute the modal policy or mode of human $m_h$ and the modal policy of robot $m_r$ towards the goal using the feature selection method described in \cite{nikolaidis2016formalizing}. $\alpha$ denotes the probability that the user will comply with robot's decision . As the adaptability is not known beforehand, it is initially assumed that the human is adaptable ($\alpha$ = 1). Based on BAM, the probability with which the user will switch to a new mode ($m_h'$) is given by:

\[ P(m_h' | \alpha, m_h, m_r)  =
\begin{cases}
    \alpha, \hspace{1cm} m_h' \equiv m_r \\
    1-\alpha, \hspace{0.5cm} m_h' \equiv m_h \\
    0, \hspace{1cm} Otherwise 
\end{cases}\]

For mutual adaptation, the robot has to estimate two variables: the human adaptability ($\alpha$) and human mode ($m_h$). Since both are not directly observable we use a mixed-observability Markov decision process (MOMDP) \cite{ong2010planning}. A reward function $R(t)$ is assigned in each step that depends on robot action $a_r$, human action $a_h$ and human mode $m_h$. The robot then maximises the function $\sum_{t=0}^{\infty} \beta^t R(t)$ where $\beta$ denotes discount factor that gives higher values to immediate rewards.

\section{Conclusion and Future Work}
In this paper, we have successfully developed a web interface for enhanced human-robot interaction. We have achieved joint-problem solving by implementing shared autonomy through the use of our web application. Further, we have described a mutual adaptation model in a shared autonomy setting to enable a robot to adapt to a human. The mutual adaption model described in the previous
section is still under construction and may be subjected to changes as the research progresses. Developing such predictive models for determining robot's decision is an exciting area of future work. A follow-up work would be to fully implement this model through our web interface and determine its usability. User studies will be carried out to compare and contrast these methods in terms of the overall performance of the system. We anticipate our web application to be an open-ended design that can be extended and built upon by other developers to use it in various research and industrial applications. 

%
%
%
\bibliographystyle{unsrt}

\bibliography{references}

\end{document}